\documentclass[manuscript,screen,nonacm]{acmart}

\AtBeginDocument{%
  \providecommand\BibTeX{{%
    \normalfont B\kern-0.5em{\scshape i\kern-0.25em b}\kern-0.8em\TeX}}}

\setcopyright{acmcopyright}
\copyrightyear{2018}
\acmYear{2020}
\acmDOI{X}

\acmJournal{TIST}
\acmVolume{X}
\acmNumber{X}
\acmArticle{X}
\acmMonth{11}

\usepackage{amsmath}
\usepackage{booktabs}
\usepackage{multirow}
\usepackage{xtab}
\usepackage{ctable}
\usepackage{array}
\usepackage{CJKutf8}
\usepackage{empheq}
\usepackage{url}
\usepackage{setspace}
\begin{document}

\title{Deep Human Answer Understanding for Natural Reverse QA}

\author{Rujing Yao}
\email{rjyao@tju.edu.cn} 
\affiliation{%
  \institution{Center for Applied Mathematics, Tianjin University}
  \streetaddress{92 Weijin Road}
  \city{Nankai Qu}
  \state{Tianjin Shi}
  \postcode{300072}
  \country{China}
}

\author{Linlin Hou}
\email{llhou@mail.nankai.edu.cn} 
\affiliation{%
  \institution{Center for Applied Mathematics, Tianjin University}
  \streetaddress{92 Weijin Road}
  \city{Nankai Qu}
  \state{Tianjin Shi}
  \postcode{300072}
  \country{China}
}

\author{Lei Yang}
\email{yl7268@tju.edu.cn} 
\affiliation{%
  \institution{Center for Applied Mathematics, Tianjin University}
  \streetaddress{92 Weijin Road}
  \city{Nankai Qu}
  \state{Tianjin Shi}
  \postcode{300072}
  \country{China}
}

\author{Jie Gui}
\email{guijie@ustc.edu} 
\affiliation{%
  \institution{Department of Computational Medicine and Bioinformatics, University of Michigan}
  \country{USA}
}

\author{Qing Yin}
\email{qingyin@tju.edu.cn} 
\affiliation{%
  \institution{Center for Applied Mathematics, Tianjin University}
  \streetaddress{92 Weijin Road}
  \city{Nankai Qu}
  \state{Tianjin Shi}
  \postcode{300072}
  \country{China}
}

\author{Ou Wu}
\email{wuou@tju.edu.cn} 
\authornote{Corresponding author}
\affiliation{%
  \institution{Center for Applied Mathematics, Tianjin University}
  \streetaddress{92 Weijin Road}
  \city{Nankai Qu}
  \state{Tianjin Shi}
  \postcode{300072}
  \country{China}
}

\renewcommand{\shortauthors}{Rujing Yao, et al.}

\begin{abstract}
This study focuses on a reverse question answering (QA) procedure, in which machines proactively raise questions and humans supply the answers. This procedure exists in many real human-machine interaction applications. However, a crucial problem in human-machine interaction is answer understanding. The existing solutions have relied on mandatory option term selection to avoid automatic answer understanding. However, these solutions have led to unnatural human-computer interaction and negatively affected user experience. To this end, the current study proposes a novel deep answer understanding network, called AntNet, for reverse QA. The network consists of three new modules, namely, skeleton attention for questions, relevance-aware representation of answers, and multi-hop based fusion. As answer understanding for reverse QA has not been explored, a new data corpus is compiled in this study. Experimental results indicate that our proposed network is significantly better than existing methods and those modified from classical natural language processing deep models. The effectiveness of the three new modules is also verified.
\end{abstract}


\maketitle

\section{Introduction}
Automatic question answering (QA) is a crucial component in many human-machine interaction systems, such as intelligent customer service, because it can provide a natural means for humans to acquire information \cite{Hixon2015,wang2018concept}. In recent years, QA has received increasing attention in academic research and industry communities \cite{Kumar2016,thukral2019diffque}. Questions are solely raised by humans, and answers are returned thereafter by machines in a conventional QA scenario such as FAQ. The manner of selecting the best matched answer is the key problem in this setting \cite{Xiong2017}.\par
\begin{figure}[htbp]
	\centering
	\includegraphics[width=0.8\linewidth]{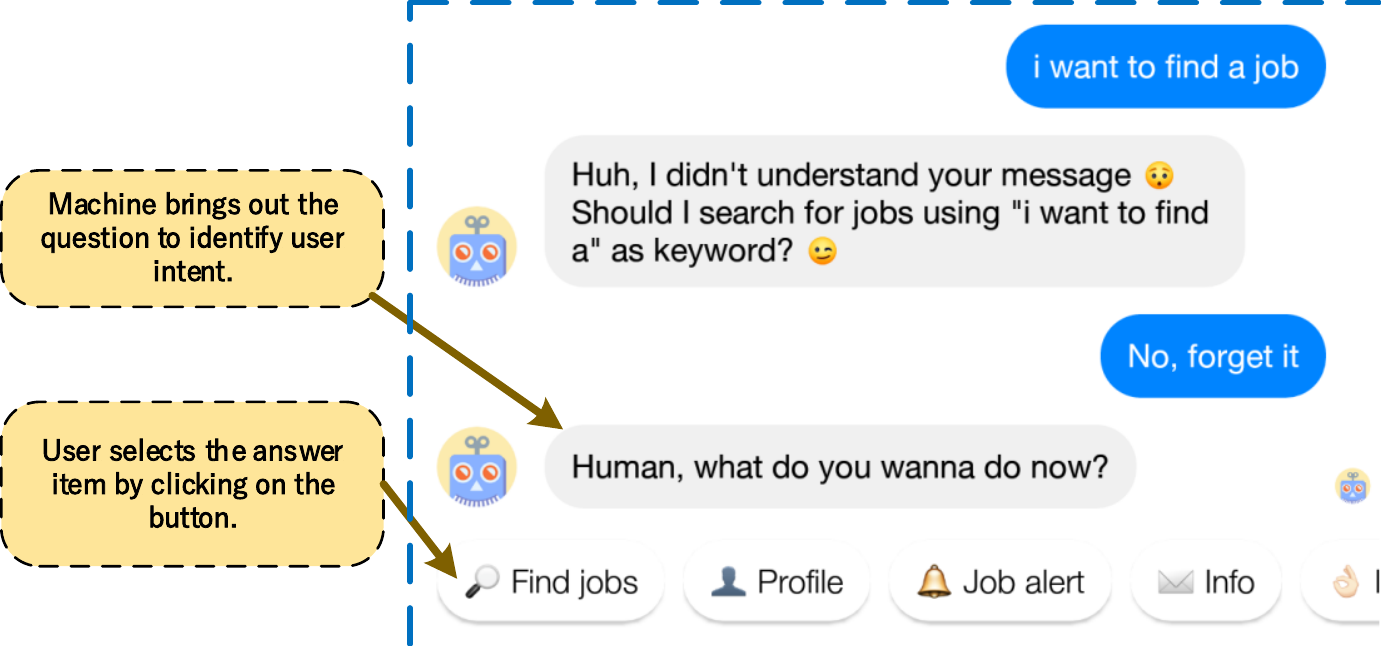}
	\caption{\label{Fig.1} Reverse QA in a commercial human-AI interaction system. Users cannot type texts for the machine questions. Instead, they are only allowed to select option items (e.g., ``Find jobs'').}
\end{figure}

Nevertheless, machines are also required to determine human needs or perceive human states in human-machine interaction systems. In such scenarios, machines proactively raise questions, and humans supply the answers. This procedure is called reverse QA. Although this process has received minimal attention in previous literature, it is common in commercial intelligent customer service systems. Fig. \ref{Fig.1} shows a reverse QA example from Facebook Job Bot\footnote{\url{https://www.facebook.com/pg/jobbot.me/about/?ref=page\_internal.}}. In nearly all commercial systems, the answer items (e.g., ``Find jobs'', ``Profile'', ``Job alert'', and ``Info'' in Fig \ref{Fig.1}) are fixed, and humans are only allowed to select at least one of the fixed candidate items. This strategy is an engineering solution, in which the interaction between user and AI system is not natural.\par
To ensure a natural human-machine interaction and improve user experience, humans should be allowed to type any texts similar to natural conversations in daily life. Moreover, machines must automatically understand the meaning of their answers without requiring them to choose fixed options shown in Fig. \ref{Fig.1}. To date, the automatic answer understanding in reverse QA has not been explored\footnote{To our knowledge, only our early work \cite{Yin2019} explored this issue. This study is an extension of our early work \cite{Yin2019}. Nevertheless, a larger data corpus is compiled and an entire new deep neural network is proposed.}.

This study proposes a new deep neural network, called answer understanding network (AntNet), on the basis of the observations on a new data corpus and inspired by the related studies, such as aspect-based sentiment analysis \cite{Chen2017, Ma2018}.\par
Given a machine-question and human-answer pair, AntNet extracts dense feature vectors for the question and the answer, and fuses the two extracted vectors thereafter. Lastly, a high-level dense feature vector is obtained and fed into a softmax layer for final answer understanding. Three new modules are included in AntNet. The first and second modules are the skeleton attention for questions and the relevance-aware representation of answers, respectively. The primary goal of the two modules is to exclude less important or even disturbing information contained in questions and answers. The third module is the multi-hop based fusion that is used to fuse answer and question vectors. Our proposed network is compared with existing methods and those with a slight modification from classical natural language processing deep models, such as Transformer \cite{Vaswani2017}.

A large data corpus\footnote{https://github.com/NlpResearch/AntNet-rverseQA} is constructed to facilitate the investigation of answer understanding in reverse QA. The experimental results indicate that AntNet is significantly better than the competing methods.\par
Our contributions are summarized as follows:
\begin{itemize}
	\item An under-explored natural language understanding task, called answer understanding in reverse QA, is investigated. To our knowledge, this is the first work that focuses on this task.
	\item A new data corpus is compiled and publicly available for interesting readers.
	\item A novel network called AntNet is proposed. This network consists of three key modules, including skeleton attention for questions, relevance-aware representation of answers, and multi-hop based fusion. AntNet significantly outperforms existing methods in the experiments.
\end{itemize}

\section{Related Work}
The most related study to reverse QA is question answering (QA). QA covers a wide range of tasks according to the application context. First, this section briefly reviews three related QA tasks, namely, text matching-based answer selection, multi-choice reading comprehension, and question generation. Second, reverse QA is reviewed and the differences between answer understanding in reverse QA and the three QA tasks are discussed.

\subsection{Text matching-based answer selection}
QA aims to return appropriate answers to users' questions. Therefore, the answers are usually selected from a corpus containing questions and answers on the basis of a text matching model in a number of studies. In some studies, the model calculates the matching scores between the question and questions in the corpus. The answers of questions with the highest matching score are then selected to return to users. Some other studies directly infer the matching score between the question and each candidate answer.

In traditional QA methods, features of questions and answers are extracted using conventional methods, such as tf-idf \cite{Salton1973}, lexical cues \cite{Khalifa2014}, and word order \cite{Hovy2001}. Thereafter, a similarity scoring function, such as cosine, is used to calculate the matching score. \par

In deep QA methods, the features of questions and answers are extracted using deep learning methods, such as convolutional neural network (CNN) \cite{Tan2016}, LSTM \cite{Tay2017}, and Transformer \cite{Vaswani2017}. An end-to-end framework is usually used to combine the deep feature extraction and successive matching function training \cite{Wang2017, Adams2018}.

Inspired by the advantage of translation in modeling the relationship between words, Xue et al. \cite{Xue20082008} used a translation-based approach to solve the problem of mismatching. Subsequently, popular neural networks like CNN \cite{Hu20142014} and LSTM \cite{Wang20152015} were used in this task. Tay et al. \cite{Tay20182018} proposed a recurrent network using temporal gates to learn the interactions between question-answer pairs.

\subsection{Multiple-choice reading comprehension}
Multiple-choice reading comprehension (MCRC) aims to select the best answer from a set of options given a question and a passage. Different from machine reading comprehension, in which the expected answer is directly contained in a given passage, answers in MCRC are non-extractive and may not appear in the original passage, thereby enabling rich types of questions, such as commonsense reasoning and passage summarization \cite{ lai2017race}.

A number of studies on MCRC model the relationship among the triplet of three sequences, namely, passage (P), question (Q) and answer (A) with a matching module to determine the answer. Zhu et al. \cite{zhu2018hierarchical} used hierarchical attention flow to explicitly model the option correlations which are ignored in previous works. Zhang et al. \cite{ zhang2019dual} leveraged the bidirectional matching strategy to gather the correlation information among the triplet \{P, Q, A\}. Thereafter, gated mechanism was introduced to fuse the representations. In the process of matching, Ran et al. \cite{ ran2019option} compared options at a word-level to effectively collect option correlation information.

\subsection{Question generation}
Many QA algorithms require labeled QA pairs as training data. Although labeled data sets, such as the WikiQA dataset \cite{Yiyang2015} for (text) QA have been proposed, these data sets are still with limited sizes because labeling is considerably expensive. This situation motivated the design of question generation to generate natural language questions from information, in which the generated questions can be answered by the contents \cite{Serban2016, Duan2017}. In this manner, large-scale QA corpus can be constructed.

Early research in question generation tackled question generation with a rule-based approach \cite{Mitkov2003} or an over generate-and-rank approach \cite{Heilman2010} which relied heavily on well-designed rules or manually crafted features, respectively. To overcome these limitations, Du et al. \cite{Du1705} introduced a deep sequence-to-sequence learning approach to generate questions. Rao et al. \cite{Rao19041904} introduced the generative adversarial networks (GANs) to generate questions that are significantly beneficial and specific to the context.

\subsection{Reverse QA}
Apart from meeting users' information requirements, machines in some real applications, such as telephone survey and commercial intelligent customer service systems, are also required to actively acquire the precise needs or feedbacks of users \cite{krcadinac2015textual}. Accordingly, machines may choose to proactively raise questions to users and then analyze their answers. That is, machines are the questioners and humans are the answerers. This process is a reverse of some text match-based QA processes (e.g., FAQ) and is called reverse QA in this study. Fig. 2 shows a conventional FAQ process and reverse QA processes.

\begin{figure}
	\centering
	\includegraphics[width=\linewidth]{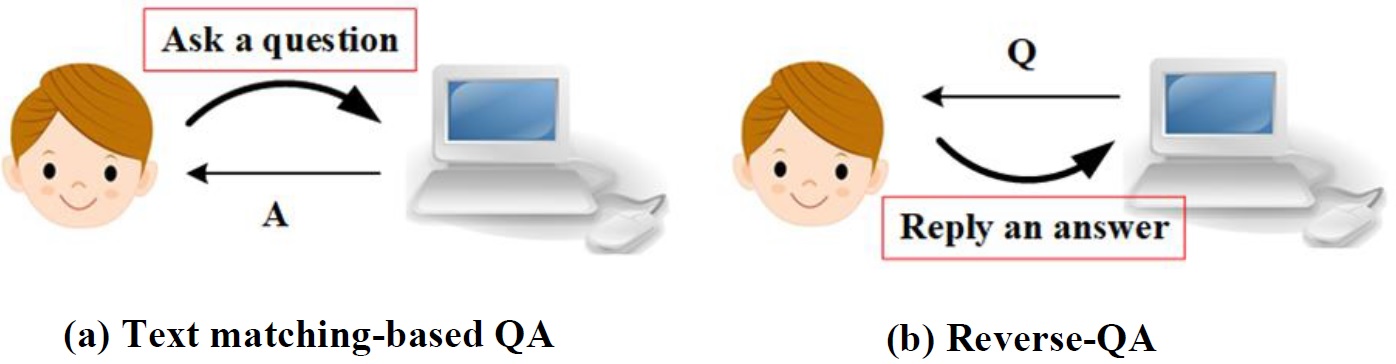}
	\caption{\label{Fig.2-0} The difference between text match-based QA (e.g., FAQ) (a) and reverse QA (b).}
\end{figure}

\begin{figure}
	\centering
	\includegraphics[width=0.5\linewidth]{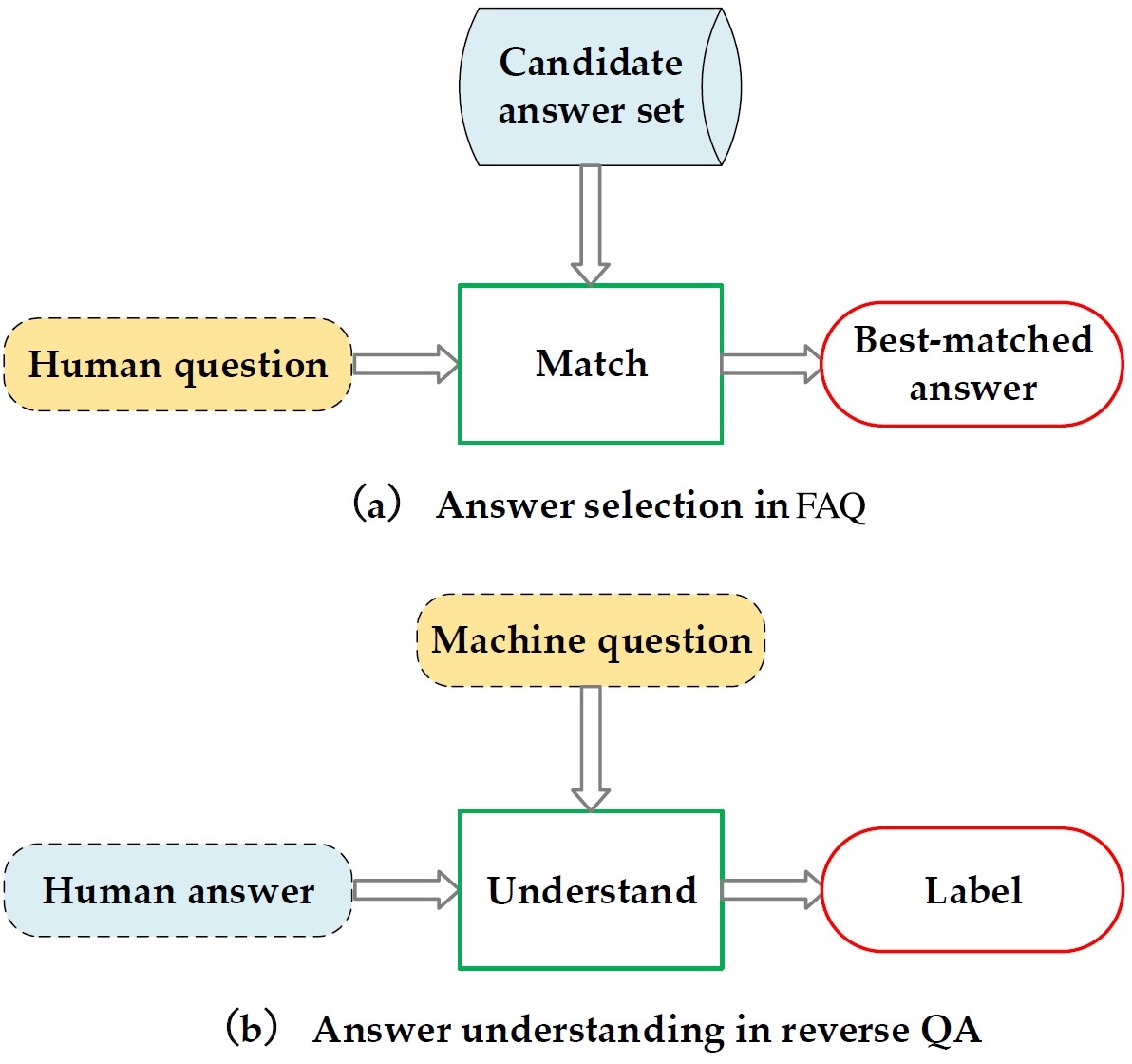}
	\caption{\label{Fig.2} The main difference between answer selection in FAQ and answer understanding in reverse QA.}
\end{figure}

The difference between answer understanding in reverse QA and text matching-based answer selection in QA is as follows. Text matching-based answer selection is a text retrieval approach, and the evaluation metrics (e.g., MAP, MRR) used for retrieval are usually applied. Consequently, the ideas of learning to rank are typically adopted. Nevertheless, the answer understanding in reverse QA is transformed into an answer classification task. Fig. \ref{Fig.2} shows the main difference between answer retrieval in text match-based answer selection in FAQ and answer classification in reverse QA\footnote{However, answer understanding investigated in this study is still a standard NLP task, so text matching can also be utilized. Our preliminary experimental results show that a simple text matching module does not improve the performance of our proposed AntNet.}.

The difference between answer understanding in reverse QA and the multi-choice reading comprehension (MCRC) in QA is as follows. Firstly, from the viewpoint of classification, MCRC is a single-label classification task, whereas answer understanding (for multi-choice questions) in this study is a multi-label classification task\footnote{Considering a machine question-human answer pair ``Q: Which day can you come here, Monday, Tuesday, or Wednesday? A: Monday or Wednesday." Both the option terms ``Monday" and ``Wednesday" are the correct answers. In MCMR, there is only one correct answer among the involved options.}. Secondly, the option texts in MCRC are independent of other inputs (i.e., paragraphs and questions), whereas the option terms in this study are contained in the questions. Text matching is the key part of MCRC.

Question generation is apparently different from answer understanding investigated in this study. Nevertheless, inspired by question generation, answer generation will be explored in our future study to alleviate the labeling load.

\section{Problem and Data}
We first provide an analysis for answer understanding in reverse QA as it is rarely investigated.
\subsection{Problem analysis }
The primary difficulty in answer understanding results from the openness of the corresponding question. For example, the three machine questions are as follows:
\begin{itemize}
	\item MQ1: Do you like sports?
	\item MQ2: Which sport do you like best, swimming, climbing, or football?
	\item MQ3: Which sport do you like?
\end{itemize}\par
MQ1 is a true/false (T/F) question, MQ2 is a multi-choice (MC) question, and MQ3 is nearly an open question. The difficulty in answer understanding for the three questions is likely to increase. The answers for MQ3 will be relatively difficult to understand considering the following answer examples: (1) ``It depends on the weather'', (2) ``Competitive sports'', and (3) ``Water sports''.\par

This study considers the T/F and MC questions. Consequently, answer understanding becomes a classification problem. The succeeding subsection presents a formal description.

%

\subsection{Problem formalization}\label{section3.2}
As previously mentioned, answer understanding for multi-choice questions can be attributed to a multi-label classification task. However, the number of categories for candidate labels for each question equals to the number of option items contained in the question. Therefore, the numbers of categories for each question are very likely to be different. The multi-label classification problem is usually transformed into one of the three existing problems, namely, binary classification, label ranking, and multi-class classification \cite{zhang2013review}. To tackle with varied numbers of label categories, the strategy of the transformation to binary classification\footnote{In fact, triple classification is actually used in this study.} is leveraged.

Let $O$ be the option term set for a question. In MC questions, $O$ is defined as the set of concrete option terms. For instance, $O$ is defined as $\{\text{\it``swimming", ``climbing", ``playing football"}\}$ for MQ2; in T/F questions, $O$ is defined as $\{\text{\it``Yes"}\}$ to ensure consistency with the format of MC questions.\par

We first illuminate how answer understanding is transformed into answer classification with concrete examples. The (answer) label set $L$ is defined as $\{\text{\it``true", ``false", ``uncertain"}\}$. Let $q_i$ be the question and $o_{i,k}$ be the $k$-th option term of $q_i$. Each question can have arbitrary numbers of answers given by users. Let $s_{i,j}$ be the $j$-th answer for $q_i$. For MQ1, given an answer $s_{i,j}$, answer understanding equals to classify $\{q_i, s_{i,j}, o_{i,k}\}$ into one of the labels in the set $L$. $o_{i,k}$ ($o_{i,k}\in O$) is $\text{\it``Yes"}$ here. For MQ2, given an answer $s_{i,j}$, answer understanding equals to three sub-classification tasks, i.e., the classification of $\{\text{\it $q_i$, $s_{i,j}$, ``swimming"}\}$, $\{\text{\it $q_i$, $s_{i,j}$, ``climbing"}\}$, and $\{\text{\it $q_i$, $s_{i,j}$, ``playing football"}\}$ into one of the labels in the label set $L$.\par

The answer classification for T/F and MC questions can be further formalized as follows:\par
{\em We aim to predict the category $l_{i,j,k}$ ($l_{i,j,k}\in L$) of the triplet $\{q_i, s_{i,j}, o_{i,k}\}$ by considering the machine-question and human-answer pair $\{q_i, s_{i,j}\}$, the corresponding option term $o_{i,k}$ of the question, and a predefined answer-label set L}.\par

The number of option terms is only one (as $O=\{\text{\it``Yes"}\}$) in T/F questions. Therefore, $o$ in the triplet can be omitted in such question type.

\subsection{Data construction}
Existing QA and text classification benchmark data sets are inappropriate for training and evaluating reverse QA models. Therefore, two data sets are compiled with a standard labeling process. The MC questions we studied are limited in the type that the options appear in the question, which we call option-contained MC questions.

For the two data sets, the questions are constructed as follows. First, seven domains are selected, namely, encyclopedia, insurance, personal, purchases, leisure interests, medical health, and exercise. A total of thirty graduate students, specifically fifteen males and fifteen females, were invited to participate in the data compiling using Email advertising from our laboratory. All the participants are Chinese and range in age from 22 to 31. Considering that the question and answer generations are not difficult to understand, we did not give special instructions to the participants. Each participant was allowed to construct 50 to 60 questions. Finally, 1543 questions are obtained after deleting some invalid questions. Among that, the numbers of T/F and MC questions are 536 and 1007, respectively.
\begin{figure*}
	\centering
	\includegraphics[width=\linewidth]{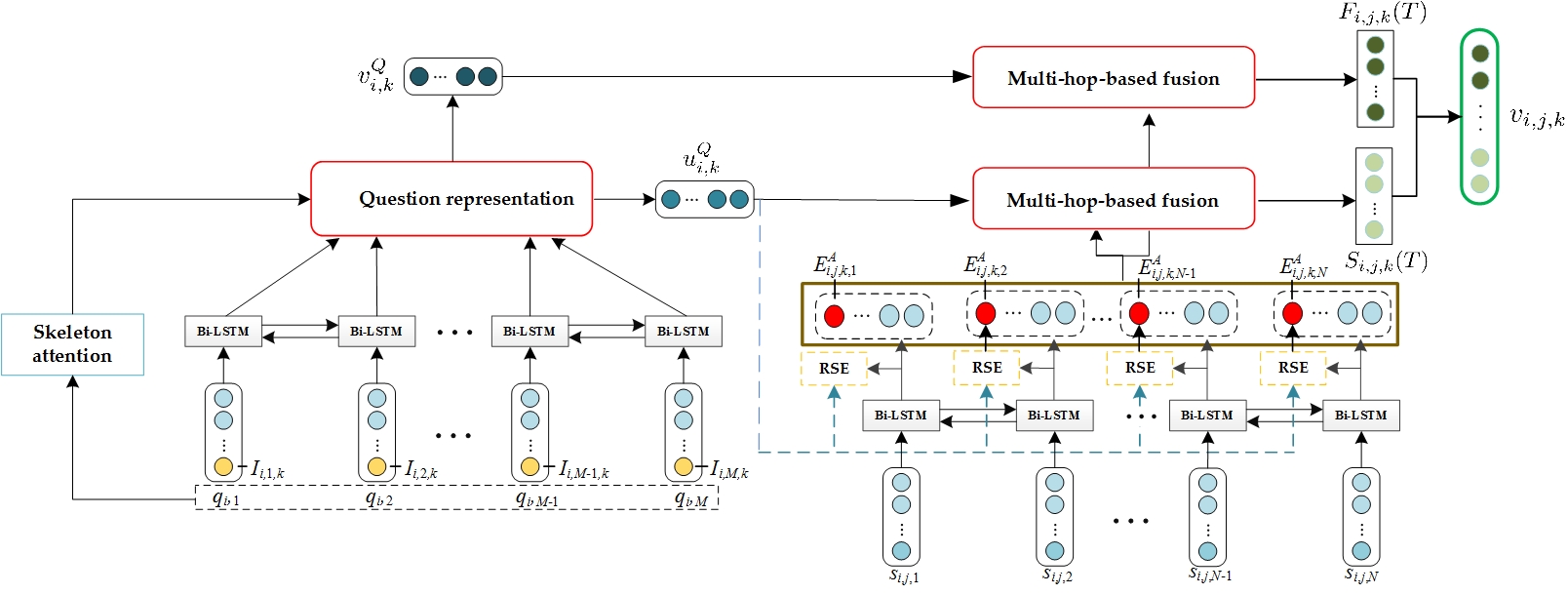}
	\caption{\label{Fig.3} The structure of AntNet.}
\end{figure*}

The questions were equally and randomly assigned to the thirty participants. Each question was given 18 to 25 answers. The participants also labeled the answers generated by themselves considering that the other participants didn't know what exactly the answer means.  A new data corpus was obtained. Table \ref{tab.1} shows the details. The data corpus contains two data sets, namely, TData and MData.
\begin{table}
	\caption{\label{tab.1}  Statistics of TData and MData}
	\centering
	\small
	\setlength{\tabcolsep}{4pt}
	\begin{tabular}{lccc}
		\toprule
		Data set               & \#Questions & \#Answers & \#Samples              \\
		                       &             &           & (True/False/Uncertain) \\
		\midrule
		\multirow{1}{*}{TData} & 536         & 10,817    & 4,610/4,452/1,755      \\

		\cmidrule(lr){1-4}
		\multirow{1}{*}{MData} & 1007        & 23,445    & 20,929/28,876/9,989    \\

		\bottomrule
	\end{tabular}

\end{table}

For the TData, the types of answers are roughly divided into affirmative, negative, uncertain, and unrelated. Given that the uncertain and unrelated answers are similar in function to the question, we classify them as the same class. Each sample consists of three components: question (i.e., $q_i$), answer (i.e., $s_{ij}$), and the associated label ($l_{ij}$) for them. The total number of samples is 10,817.

For the MData, the numbers of option terms for each MC question are different and cannot be categorized uniformly. Thus, we add the option information to the MC questions and get a series of transformed MC questions as described in
Section \ref{section3.2}. Therefore, the same answer to the same question will have different labels for dissimilar option terms. Each sample consists of four components: question, option, answer, and label. There are 59,794 transformed MC training samples.

\section{Methodology}
Section \ref{section3.2} describes that answer understanding is transformed into an answer classification problem. The first step is obtaining the deep representations of the machine-question and human-answer pair and a given option term. In addition, questions provide the context for answer understanding. The final dense representation should consider the contextual dependency between questions and answers.\par

The related research in text classification and aspect-based sentiment analysis inspired us to propose a new deep model called AntNet. Fig. \ref{Fig.3} shows the main structure of this model.\par The experimental data are in Chinese. Therefore, the word means ``the Chinese word" in the following subsections.

The AntNet input is the triplet $\{q_i, s_{i,j}, o_{i,k}\}$, where $o_{i,k}$ is indicated by an indicator vector. The indicator is set as a zero vector for all samples in T/F questions, and the option indicator is set as a one-hot vector in MC questions. The left part of AntNet deals with the input of $q_i$ and $o_{i,k}$ to generate two representations. The first representation characterizes the combination of $q_i$ and $o_{i,k}$, while the second representation characterizes important information, which is called skeleton (Chinese) words for questions in this study. The first and second representations are called full and skeleton representations, respectively.\par
The lower-right portion deals with the input of answers, and the output is a set of hidden dense vectors for answers. In this part, a relevance-aware module is used to substantially characterize the relevance cues contained in the answers, which consider that users may return irrelevant texts.\par
The upper-right portion deals with the contextual dependency between questions and answers to obtain an overall dense feature vector, which is fed into the final decision softmax layer. A multi-hop attention mechanism is used in this part.\par
The following subsections introduce the details of the three parts. The skeleton attention is firstly introduced.

\subsection{Skeleton attention}
Question texts usually contain redundant\footnote{These words may be used for enhance the interestingness of the interaction.} or even disturbed words, which may negatively influence answer understanding. The skeleton information in a question should be extracted. Skeleton information refers to words that directly affect how users respond to.

Intuitively, skeleton information extraction can be performed in a supervised manner. Alternatively, skeleton words are manually labeled for a set of training text samples. Thereafter, these training samples are fed into a sequence labeling model for training. The trained sequence labeling model can be used to extract skeleton words for new texts. Nevertheless, providing an explicit and formal definition for skeleton words is difficult, thereby making it difficult for human labeling as well. To this end, this study proposes an attention-based manner.

In this study, a training sample is a triplet $\{q_i, s_{i,j}, o_{i,k}\}$, where $q_i$ is the $i$-th question, $s_{i,j}$ is the $j$-th answer for $q_i$, and $o_{i,k}$ is the $k$-th option term for $q_i$. The primary difference between the current study and conventional classification studies lies in that many training samples in this study share the same element `$q_i$'. That is, each question ($q_i$) corresponds to multiple answers ({$s_{i,1}, \cdots, s_{i,j}, \cdots, s_{i,J}$}), leading that there are multiple training samples for $q_i$ and a fixed option term $o_{i,k}$ including $\{q_i, s_{i,1}, o_{i,k}\}, \cdots, \{q_i, s_{i,J}, o_{i,k}\}$. Let $q_i = \{q_{i,1}, \cdots, q_{i,m}, \cdots, q_{i,M_i}\}$ be the $i$-th question, where $M_i$ is the word-level length of the question and $q_{i,m}$ is the $m$-th word of $q_i$. Let $s_{i,j} = \{s_{i,j,1}, \cdots, s_{i,j,n}, \cdots, s_{i,j,N_{ij}}\}$ be the $j$-th answer for $q_i$, where $N_{ij}$ is the word-level length of the answer and $s_{i,j,n}$ is the $n$-th word. An attention score can be calculated for $q_{i,m}$ as follows.
\begin{equation}
	\omega (q_{i,m}) = \frac{1}{J\cdot N_{ij}}\sum\limits_{j = 1}^J {\sum\limits_{n = 1}^{N_{ij}} {{\emph{sim}}}(q_{i,m},{s_{i,j,n}}){\rm{ }}}
\end{equation}
where $sim(q_{i,m},{s_{i,j,n}}) = q_{i,m}^T W_s{s_{i,j,n}}$ calculates the similarity of two words according to their word embeddings and the matrix $W_s$ are learned during training.

\begin{figure}
	\centering
	\includegraphics[width=1\textwidth]{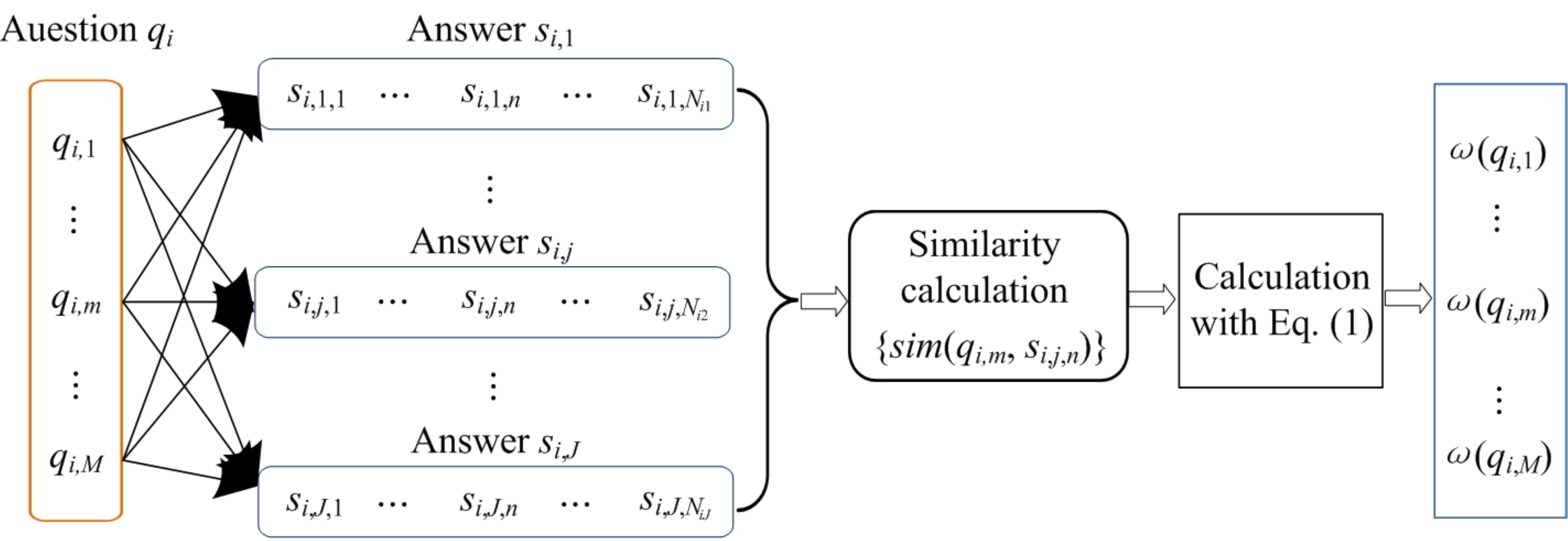}
	\caption{\label{Fig.3-1} Skeleton attention score calculation pipeline.}
\end{figure}

\begin{figure}
	\centering
	\includegraphics[width=0.6\textwidth]{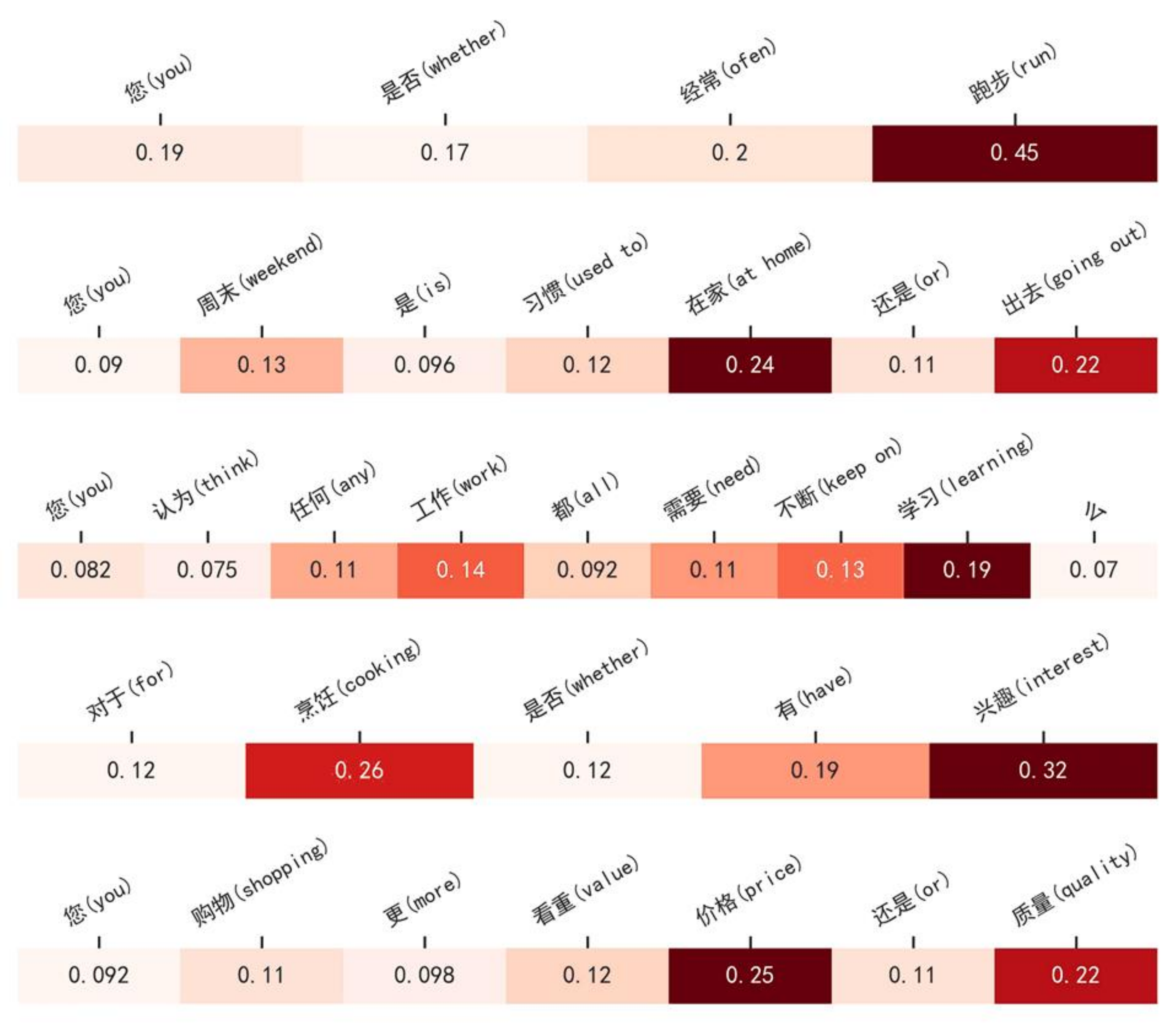}
	\caption{\label{Fig.3-3} Attention scores for words in five questions. All the data in this study are in Chinese. To facilitate English readers, the Chinese words in the above questions are translated into English.}
\end{figure}

Thereafter, the scores calculated by Eq. (1) are then normalized as attention scores. Fig. \ref{Fig.3-3} shows several questions and their associated attention scores on each Chinese word calculated using Eq. (1). The words with higher scores are key words in their corresponding questions. The scores of such words as ``you", ``are", ``this", and ``or" are low in most sentences. The words that are directly related to user options, such as ``run", ``interest", ``quality", and ``at home" have high scores.

\subsection{Question representation}
AntNet considers two-level representations. The first-level representation (referred to skeleton representation) characterizes the skeleton information in the question, while the second-level representation (referred to full representation) characterizes the entire question. The two representation vectors are calculated as follows.\par
The given training sample is represented by an input triplet $\{q_i, s_{i,j}, o_{i,k}\}$ and its label $l_{i,j,k}$. Let $I_{i,m,k}$ be an indication vector for whether the word $q_{i,m}$ is in $o_{i,k}$\footnote{Some option terms are phrases.}. If $q_{i,m}$ is in $o_{i,k}$, then $I_{i,m,k}=1$; otherwise $I_{i,m,k}=0$.\par

After the encoding of BiLSTM on $q_i$, the hidden representation of each word is defined as follows:
\begin{eqnarray}
	h_{i,m,k}^Q = BiLSTM_Q(h_{i,m-1,k}^Q, h_{i,m+1,k}^Q, q_{i,m},I_{i,m,k})
\end{eqnarray}
where $h_{i,m,k}^Q \in \mathbb{R}^d$ and $q_{i,m}$ represent the embedding vector of words.\par

Given that the skeleton attention is calculated using Eq. (1), the skeleton representation for a question $q_i$ (given $o_{i,k}$) is calculated as follows:
\begin{eqnarray}
	{u_{i,k}^Q} = \sum\limits_{q_{i,m} \in {Sk_{i}}} {\omega (q_{i,m})h_{i,m,k}^Q/\sum\limits_{q_{i,m} \in {Sk_{i}}}{\omega (q_{i,m})}}.
\end{eqnarray}

The full representation $v_{i,k}^Q$ of $q_i$ (given the involved option term $o_{i,k}$) is calculated on the basis of attention scores $\{{att_{i,m,k}^Q}\}_{m=1}^{M_i}$ for each word $q_{i,m}$. The calculation is described as follows:
\begin{eqnarray}\label{5,6,7}
	&a_{i,m,k}^Q={h_{i,m,k}^Q}^TW_au_{i,k}^Q\nonumber\\
	&att_{i,m,k}^Q=\frac{exp(a_{i,m,k}^Q)}{\sum_{m=1}^{M_i}exp(a_{i,m,k}^Q)}\\
	&v_{i,k}^Q=\sum_{m=1}^{M_i}att_{i,m,k}^Qh_{i,m,k}^Q\nonumber
\end{eqnarray}
where $W_a\in \mathbb{R}^{d\times d}$, $a_{i,m,k}^Q, att_{i,m,k}^Q \in \mathbb{R}$, and $v_{i,k}^Q \in \mathbb{R}^d$.
\subsection{Relevance-aware answer representation}
BiLSTM is also utilized to generate the hidden vectors of answer texts with the following calculation:
\begin{eqnarray}\label{8}
	h_{i,j,n}^A = BiLSTM_A(h_{i,j,n-1}^A, h_{i,j,n+1}^A, s_{i,j,n})
\end{eqnarray}
where $h_{i,j,n}^A \in \mathbb{R}^d$.

To maintain the naturalness of the entire interaction, users can return their answers in arbitrary forms and with arbitrary contents. Therefore, some irrelevant texts are included in some answers even if these answers do not belong to the ``irrelevant" category. Thereafter, a relevance score is calculated for each word ($s_{i,j,n}$) in the answer texts ($s_{i,j}$) with the following equation:
\begin{eqnarray}\label{9}
	p_{i,j,k,n}^A=sigmoid(W_p[h_{i,j,n}^A,u_{i,k}^Q]+b_p).
\end{eqnarray}

\begin{figure}[htbp]
	\centering
	\includegraphics[width=0.7\textwidth]{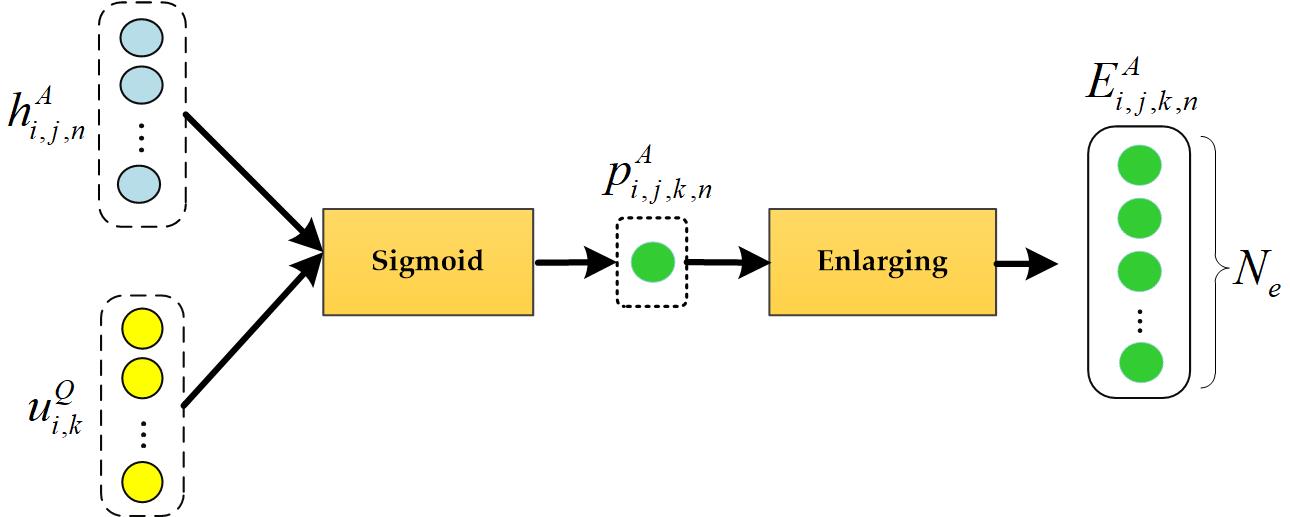}
	\caption{\label{Fig.preference} The \textbf{r}elevance \textbf{s}core calculation and dimension \textbf{e}nlarging (RSE).}
\end{figure}

The length of $p_{i,j,k,n}^A$ is one which is substantially smaller than $h_{i,j,n}^A$ in our practical implementation. Consequently, the proportion of the $p_{i,j,n}^A$ part is relatively small in the concatenated vectors, thereby limiting the advantages of the relevance vectors. We adopt the trick used in \cite{Wu2018} in our implementation. The length of $p_{i,j,n}^A$ is enlarged as follows:
\begin{eqnarray}\label{11}
	E_{i,j,k,n}^A=p_{i,j,k,n}^A\otimes 1_{N_e\times 1}
\end{eqnarray}
where $1_{N_e\times 1}$ is an $N_e$-dimensional vector. $E_{i,j,k,n}^A$ is the enlarged vector, and parameter $N_e$ is used to increase the length of $p_{i,j,k,n}^A$. Fig. \ref{Fig.preference} shows the steps of  the \textbf{r}elevance \textbf{s}core calculation and dimensionality \textbf{e}nlarging (RSE). Experimental results validate the effectiveness of the dimensionality increment for $p_{i,j,k,n}^A$.\par
The relevance score vector is concatenated with the hidden vectors for each word as follows:
\begin{eqnarray}\label{10}
	{h'}_{i,j,k,n}^A=\left[
		\begin{array}{cc}  
			h_{i,j,n}^A \\
			E_{i,j,k,n}^A
		\end{array}
		\right ]                              
\end{eqnarray}
where ${h'}_{i,j,k,n}^A\in \mathbb{R}^{d+N_e}$ is the updated hidden representation of each answer word.\par

\subsection{Multi-hop based fusion}
The representations (i.e., $u_{i,k}^Q, v_{i,k}^Q, {h'}_{i,j,k}^A$) are fused to obtain the final representation of the entire triplet $\{q_i, s_{i,j}, o_{i,k}\}$.\par
Inspired by the ABSA \cite{Tang2016}, a multi-hop based question-answer fusion module is introduced. This module can substantially represent the input machine-question and human-answer pair and the associated option term.\par
\begin{figure}[htbp]
	\centering
	\includegraphics[width=0.9\textwidth]{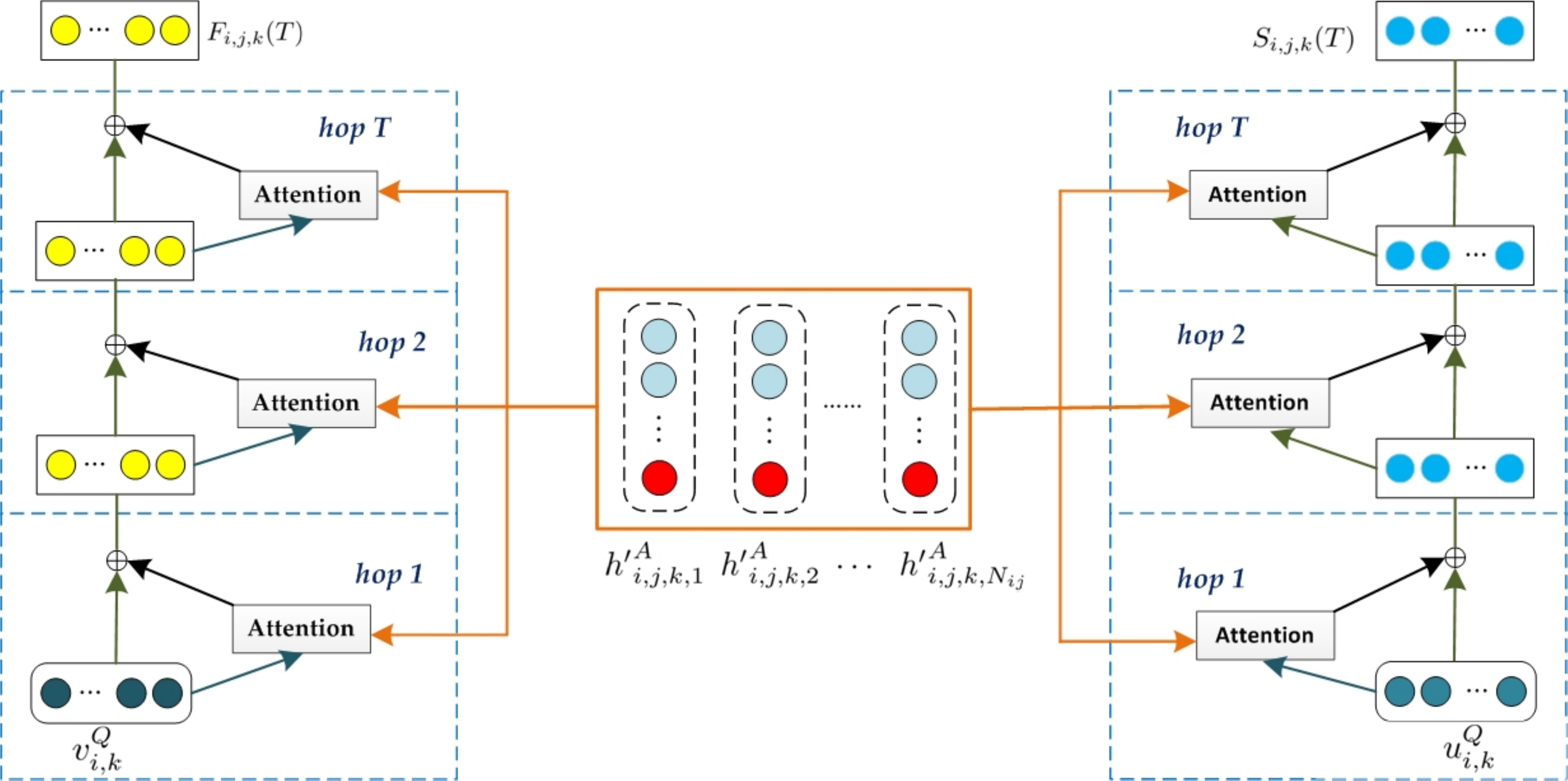}
	\caption{\label{Fig.4} Multi-hop based fusion for the involved question (two feature vectors $u_{i,k}^Q$ and $v_{i,k}^Q$) and answer (${h'}_{i,j,k}^A$).}
\end{figure}

The vectors $u_{i,k}^Q$ and $v_{i,k}^Q$ are seperately input into the multi-hop based fusion module. Fig. \ref{Fig.4} shows the multi-hop based fusion. The left part and the right part are the iterative approaches for $v_{i,k}^Q,u_{i,k}^Q$, respectively, given ${h'}_{i,j,k}^A$.
\par
The calculation with $v_{i,k}^Q$ and ${h'}_{i,j,k}^A$ is used as an example. Let $F_{i,j,k}(0) = v_{i,k}^Q$ be the input question representation. The first hop (hop 1 in Fig. \ref{Fig.4}) is computed as follows:
\begin{eqnarray}\label{12,13,14,15}
	&F_{i,j,k}(0)=v_{i,k}^Q\nonumber\\
	&m_{i,j,k,n}^{(1)}=W_m^{(1)}tahn(W_h^{(1)}{h'}_{i,j,k,n}^A+W_x^{(1)}F_{i,j,k}(0)+b^{(1)})\nonumber\\
	&a_n^{(1)}=\frac{exp(m_{i,j,k,n}^{(1)})}{\sum_{n=1}^{N_{i,j}}exp(m_{i,j,k,n}^{(1)})}\\
	&x'=\sum_{n=1}^{N_{i,j}}a_n^{(1)}{h'}_{i,j,k,n}^A\nonumber.
\end{eqnarray}\par
An active module is used to obtain the following new vector:
\begin{eqnarray}\label{16}
	F_{i,j,k}(1)=tanh(W_{f1}x'+b_f)+W_{f2}F_{i,j,k}(0)
\end{eqnarray}
where $F_{i,j,k}(1)$ is also the input of the second hop (hop 2 in Fig. \ref{Fig.4}).\par
The preceding step is iterated $T$ times to obtain the feature vector $F_{i,j,k}(T)$.\par
Lastly, $F_{i,j,k}(T)$ from the full representation $v_{i,k}^Q$ and $S_{i,j,k}(T)$ from the skeleton question representation $u_{i,k}^Q$ are concatenated into one representation vector:
\begin{eqnarray}\label{17}
	v_{i,j,k}=\left[
		\begin{array}{ccc}  
			F_{i,j,k}(T) \\
			S_{i,j,k}(T)
		\end{array}
		\right ].                        
\end{eqnarray}\par
The predicted label is calculated as follows:
\begin{eqnarray}\label{18}
	l'_{i,j,k}=softmax(Wv_{i,j,k}+b).
\end{eqnarray}\par
Given the predicted and ground truth labels, AntNet can be learned with the following cross-entropy loss function:
\begin{eqnarray}\label{19}
	loss=-\sum_{i,j,k}l_{i,j,k}logl_{i,j,k}'.
\end{eqnarray}

\section{Experiments}
This section presents the evaluation of the proposed AntNet in terms of the entire network and the three key modules, namely, skeleton attention for questions, relevance-aware representation of answers and multi-hop based fusion.
\subsection{Competing methods}
Several classical and state-of-the-art deep model-based algorithms are used and listed as follows:
\begin{itemize}
	\item \textbf{BiLSTM (A)}: Standard BiLSTM is used to encode the answer texts directly, and the dense vector is used for answer classification.
	\item \textbf{BiLSTM (Q+A)}: Standard BiLSTM is also used for the question and answer texts.
	\item \textbf{RAM} \cite{Chen2017}: RAM leverages the hidden vectors of BiLSTM as memory vectors. Then, GRU is used to construct a multi-hop based fusion for memory and input target vectors. The final dense vector contains information from sentences and targets. This study takes question texts as target texts.
	\item \textbf{ATAE} \cite{Wang2016}: ATAE is based on BiLSTM and proposed for target-based sentiment analysis. The target vector is concatenated with the word embedding of each word. In this experiment, the question texts are taken as the target texts.
	\item \textbf{Transformer (A)}: The standard Transformer is used to encode the answer texts directly and the averaging pooling of the hidden vectors of the last layer is used for answer classification. BERT \cite{DBLP:journals/corr/abs-1810-04805} is used to infer the embedding vectors for each word.
	\item \textbf{Transformer (Q+A)}: Questions and answers are concatenated and input into the standard transformer with BERT.
	\item \textbf{Semi-IAN} \cite{Yin2019}: Semi-IAN is our early proposed network related to answer understanding in reverse QA. The interaction between question and answers are modeled. Semi-IAN is based on an ABSA network called interactive attention network (IAN) \cite{Tang2016}.
	\item \textbf{(Python) Regularized Matching (RM)}: This method is an engineering solution that matches pre-defined key words or phrases, or their combinations.
\end{itemize}\par
Our proposed method consists of several new modules. To investigate the validity of three major components, namely, skeleton attention, relevance-aware answer representation, and multi-hop based fusion, we test AntNet with or without these components. The variants of our method are listed as follows:
\begin{itemize}
	\item \textbf{AntNet}: The entire AntNet with all introduced key components.
	\item \textbf{AntNet$-$SA}: The AntNet without the skeleton attention.
	\item \textbf{AntNet$-$RR}: The AntNet without the relevance-aware representation.
	\item \textbf{AntNet$-$MF}: The AntNet without the multi-hop based fusion.
	\item \textbf{AntNet$-$SA$-$RR}: The AntNet without the skeleton attention and the relevance-aware representation.
	\item \textbf{AntNet$-$RR$-$MF}: The AntNet without the relevance-attention representation and the multi-hop based fusion.
	\item \textbf{AntNet$-$MF$-$SA}: The AntNet without the multi-hop based fusion and the skeleton attention.
\end{itemize}\par
Given that the answer understanding for reverse QA is investigated from a classification perspective, the classification accuracy and F1 score are used as the performance metrics.
\subsection{Training setting}
Two data corpora, namely, TData and MData, are involved in our experiment. They are divided according to the following rules:\par
(1)	Each data corpus is divided into two parts with the 4:1 proportion. Four folds are used for training, and the remainder is used for testing.\par
(2)	A proportion of 10\% samples in training data are used as validation data.\par
We use 256-dimension Word2Vector embeddings trained on our own corpus. We initialize the word embeddings randomly for the out-of-vocabulary words. In addition, the lengths of questions and answers are truncated as 33. The learning rate and dropout ratio are set to $5\times10^{-4}$ and 0.2, respectively. We minimize the loss function using the ADAM optimizer \cite{Kingma2015} and fix the word embedding vectors during the training. Furthermore, we set the remaining parameters as default values.\par
All the aforementioned models are trained using Tensorflow.

\subsection{Overall competing results}
Table II presents the main results (classification accuracies and F1 scores) of the competing methods on the two data corpora. AntNet achieves the highest accuracies on both data corpora. Compared with the state-of-the-art network, Transformer, the results significantly increased. The relatively poor performance of Transformer may result from the small training size.\par

\begin{table}[htbp]
	\caption{\label{tab.3} Accuracies on TData and MData.}
	\centering
	\small
	\setlength{\tabcolsep}{4.1pt}
	\begin{tabular}{lcccc}
		\toprule
		\multirow{2}*{Method}              & \multicolumn{2}{c}{TData} & \multicolumn{2}{c}{MData}                                         \\
		\cline{2-5}
		~                                  & Accuracy                  & F1                        & Accuracy          & F1                \\
		\midrule
		\multirow{1}{*}{BiLSTM (A)}        & 0.7375                    & 0.7196                    & 0.6701            & 0.6681            \\
		\multirow{1}{*}{BiLSTM (Q+A)}      & 0.7196                    & 0.6982                    & 0.6738            & 0.6868            \\
		\multirow{1}{*}{RAM}               & 0.7503                    & 0.7435                    & 0.7036            & 0.6860            \\
		\multirow{1}{*}{ATAE}              & 0.7458                    & 0.7361                    & 0.7064            & 0.7446            \\
		\multirow{1}{*}{Transformer (A)}   & 0.7435                    & 0.7343                    & 0.6741            & 0.6525            \\
		\multirow{1}{*}{Transformer (Q+A)} & 0.7167                    & 0.6911                    & 0.6966            & 0.6537            \\
		\multirow{1}{*}{Semi-IAN}          & 0.7485                    & 0.7427                    & 0.7086            & 0.6871            \\
		\multirow{1}{*}{AntNet}            & $\textbf{0.7986}$         & $\textbf{0.7923}$         & $\textbf{0.8419}$ & $\textbf{0.8517}$ \\
		\bottomrule
	\end{tabular}
\end{table}

\begin{table}[htbp]
	\caption{\label{tab.2} Accuracies on TData with DuReader YES\_NO Data.}
	\centering
	\small
	\setlength{\tabcolsep}{5pt}
	\begin{tabular}{lcc}
		\toprule
		Method                             & Accuracy          & F1                \\
		\midrule
		\multirow{1}{*}{BiLSTM (A)}        & 0.7445            & 0.7334            \\
		\multirow{1}{*}{BiLSTM (Q+A)}      & 0.7454            & 0.7423            \\
		\multirow{1}{*}{RAM}               & 0.7701            & 0.7674            \\
		\multirow{1}{*}{ATAE}              & 0.7492            & 0.7201            \\
		\multirow{1}{*}{Transformer (A)}   & 0.7529            & 0.7436            \\
		\multirow{1}{*}{Transformer (Q+A)} & 0.7267            & 0.7126            \\
		\multirow{1}{*}{Semi-IAN}          & 0.7523            & 0.7485            \\
		\multirow{1}{*}{AntNet}            & $\textbf{0.8045}$ & $\textbf{0.8048}$ \\
		\bottomrule
	\end{tabular}
\end{table}

The existing answer understanding method (i.e., Semi-IAN) is inferior to RAM. In fact, Semi-IAN is a slight variation of the ABSA network IAN. Given that RAM is also an ABSA method, RAM unsurprisingly outperforms Semi-IAN. Among these methods, the RM method has the lowest accuracy of 50.38\% on average. Hence, a machine learning-based approach is essential.

An existing QA corpus DuReader \cite{He37} was used to pre-train the involved models. DuReader\footnote{http://ai.baidu.com/broad/download?dataset=dureader} is a large-scale real-world Chinese dataset in which there are three types of questions, namely, `DESCRIPTION', `ENTITY' and `YES\_NO'. The `YES\_NO' samples contain the information required for T/F questions but the option term is missing for MC questions. Therefore, the corpus can only be used for pre-training for T/F questions. Table III shows the results. The performances of all the competing methods are improved, although the increase is not significant.

\subsection{Evaluation of the different modules of AntNet}
This subsection verifies the usefulness of the three introduced key modules, namely, skeleton representation of questions, relevance-aware representation of answers, and multi-hop based fusion. The involved competing methods are AntNet$-$SA, AntNet$-$RR, AntNet$-$MF, AntNet$-$SA$-$RR, AntNet$-$SA$-$MF, AntNet$-$RR$-$MF, and the entire network AntNet.\par

Table IV shows the competing results on the two data corpora, TData and MData. Unsurprisingly, all variations without a certain type of key module achieve inferior accuracies compared with the full version of AntNet. The performances of the variations without two key modules decreases heavily. These comparisons indicate that the three key modules are benefical in answer understanding.\par

\begin{table}
	\caption{Results of AntNet and its variations (without certain key modules) on TData and MData.}
	\centering
	\small
	\setlength{\tabcolsep}{4.8pt}
	\begin{tabular}{lcccc}
		\toprule
		\multirow{2}*{Method}                                & \multicolumn{2}{c}{TData} & \multicolumn{2}{c}{MData}                                         \\
		\cline{2-5}
		~                                                    & Accuracy                  & F1                        & Accuracy          & F1                \\
		\midrule
		\multirow{1}{*}{AntNet}                              & $\textbf{0.8045}$         & $\textbf{0.8048}$         & $\textbf{0.8419}$ & $\textbf{0.8517}$ \\
		\multirow{1}{*}{\space\space\space\space $-$SA}      & 0.7863                    & 0.7774                    & 0.8238            & 0.8374            \\
		\multirow{1}{*}{\space\space\space\space $-$RR}      & 0.7907                    & 0.7790                    & 0.7107            & 0.7094            \\
		\multirow{1}{*}{\space\space\space\space $-$MF}      & 0.7740                    & 0.7551                    & 0.7050            & 0.6897            \\

		\multirow{1}{*}{\space\space\space\space $-$SA$-$RR} & 0.7760                    & 0.7681                    & 0.7033            & 0.7026            \\
		\multirow{1}{*}{\space\space\space\space $-$SA$-$MF} & 0.7686                    & 0.7494                    & 0.6809            & 0.6736            \\
		\multirow{1}{*}{\space\space\space\space $-$RR$-$MF} & 0.7714                    & 0.7523                    & 0.6943            & 0.6915            \\
		\bottomrule
	\end{tabular}
\end{table}

\newcommand{\tabincell}[2]{\begin{tabular}{@{}#1@{}}#2\end{tabular}}
\begin{figure}[htbp]
	\centering
	\includegraphics[width=0.6\textwidth]{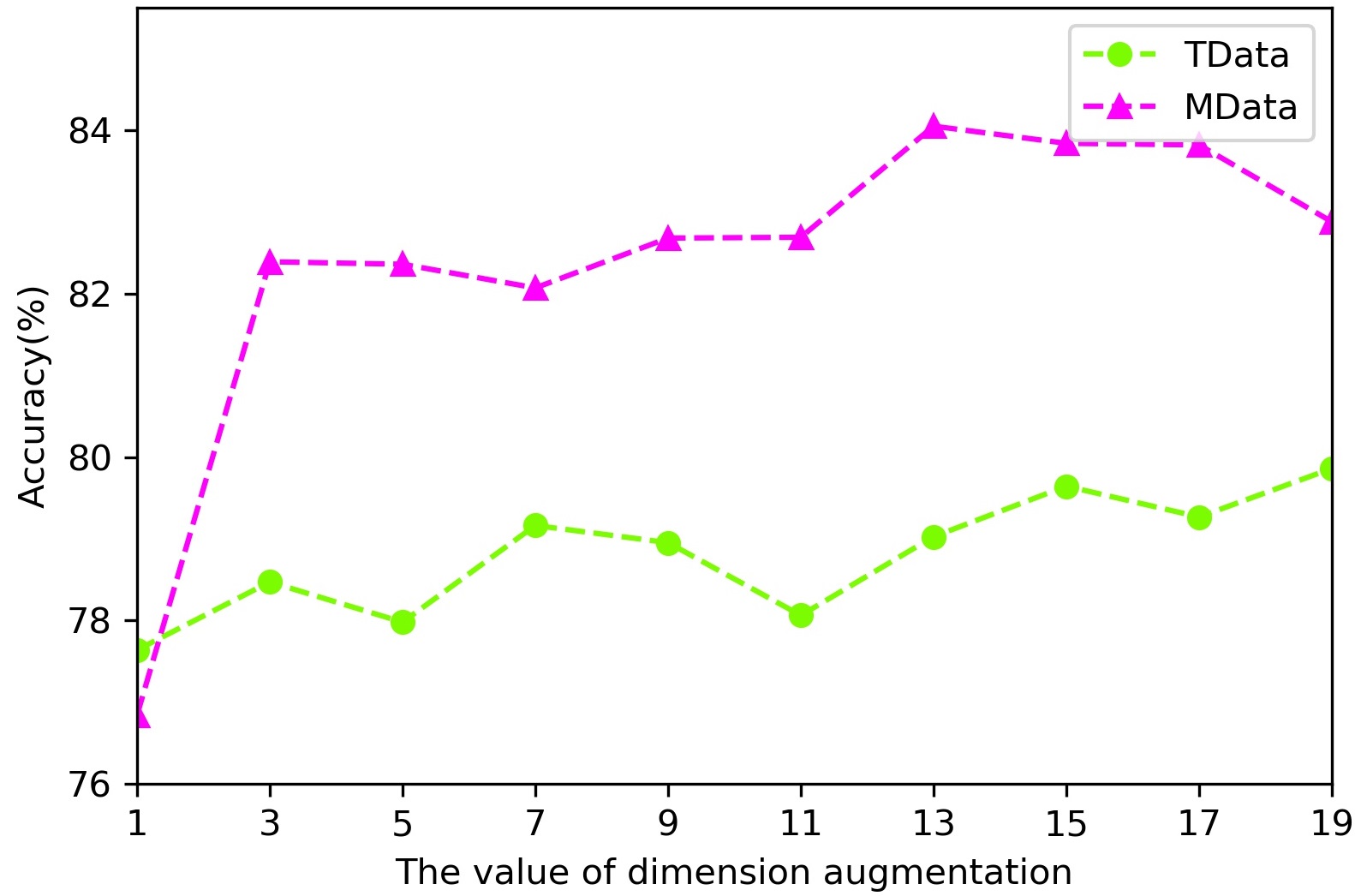}
	\caption{\label{Fig.5} Understanding accuracies under the different values of dimension augmentation for relevance-aware representation.}
\end{figure}\par

\begin{CJK*}{UTF8}{gkai}
\begin{table*}[t]
	\caption{\label{tab.6} Examples in which human answers contain implicit relevance hints.}
	\begin{center}
		\small
		\begin{tabular}{|c|c|c|}
			\hline \tabincell{c}{  \\\bf Machine question\\ \\ } & \tabincell{c}{ \\ \bf Human answer\\  \\}  & \tabincell{c}{ \\ \bf Labels \\ \\} \\ \hline
			\tabincell{c}{你是喜欢裙子还是裤子?                               \\(Do you like skirts or pants?)}  & \tabincell{c}{我不挑.\\(I am not picky.)} &  \tabincell{c}{``True" for both ``skirts" and ``pants".} \\ \hline
			\tabincell{c}{您周末喜欢逛街还是打游戏?                           \\(Do you like shopping or playing \\games on weekends?)}    &  \tabincell{c}{我是女生哎!\\(I'm a girl.)}&  \tabincell{c}{``True" for ``shopping" and \\``False" for ``playing games".}\\ \hline
			\tabincell{c}{您平时喜欢喝热水还是凉水?                           \\(Do you like hot or cold water?)}    &  \tabincell{c}{我爱喝苏打水.\\(I like to drink soda.)}&   \tabincell{c}{``False" for both ``hot" and ``cold water".} \\ \hline
			\tabincell{c}{您习惯晨跑还是夜跑?                                 \\(Are you used to running in the \\morning or at night?) }&  \tabincell{c}{我喜欢看别人跑.\\ (I like to watch others run.)}& \tabincell{c}{``False" for both ``in \\the morning" and ``at night".}\\ \hline
		\end{tabular}
	\end{center}
\end{table*}
\end{CJK*}

The comparison of the three variations shows that AntNet$-$SA$-$MF (AntNet without SA and MF) obtains the lowest accuracies. On MData, the accuracy achieved by AntNet$-$SA$-$MF is approximately 16.1\% lower than that by AntNet.

In the relevance-aware representation, the dimension of the relevance score is augmented by using Eq. (\ref{11}). We perform an experiment to investigate the performances of AntNet under different augment parameters $N_e$ in Eq. (\ref{11}). Fig. \ref{Fig.5} shows the accuracies of AntNet according to different $N_e$ values. With the increase of the value of $N_e$, the understanding accuracies on both sets demonstrate an increasing trend. When the values equal 13 and 19, AntNet achieves the maximum accuracies on both data sets.\par

\begin{figure}[htbp]
	\centering
	\includegraphics[width=0.6\textwidth]{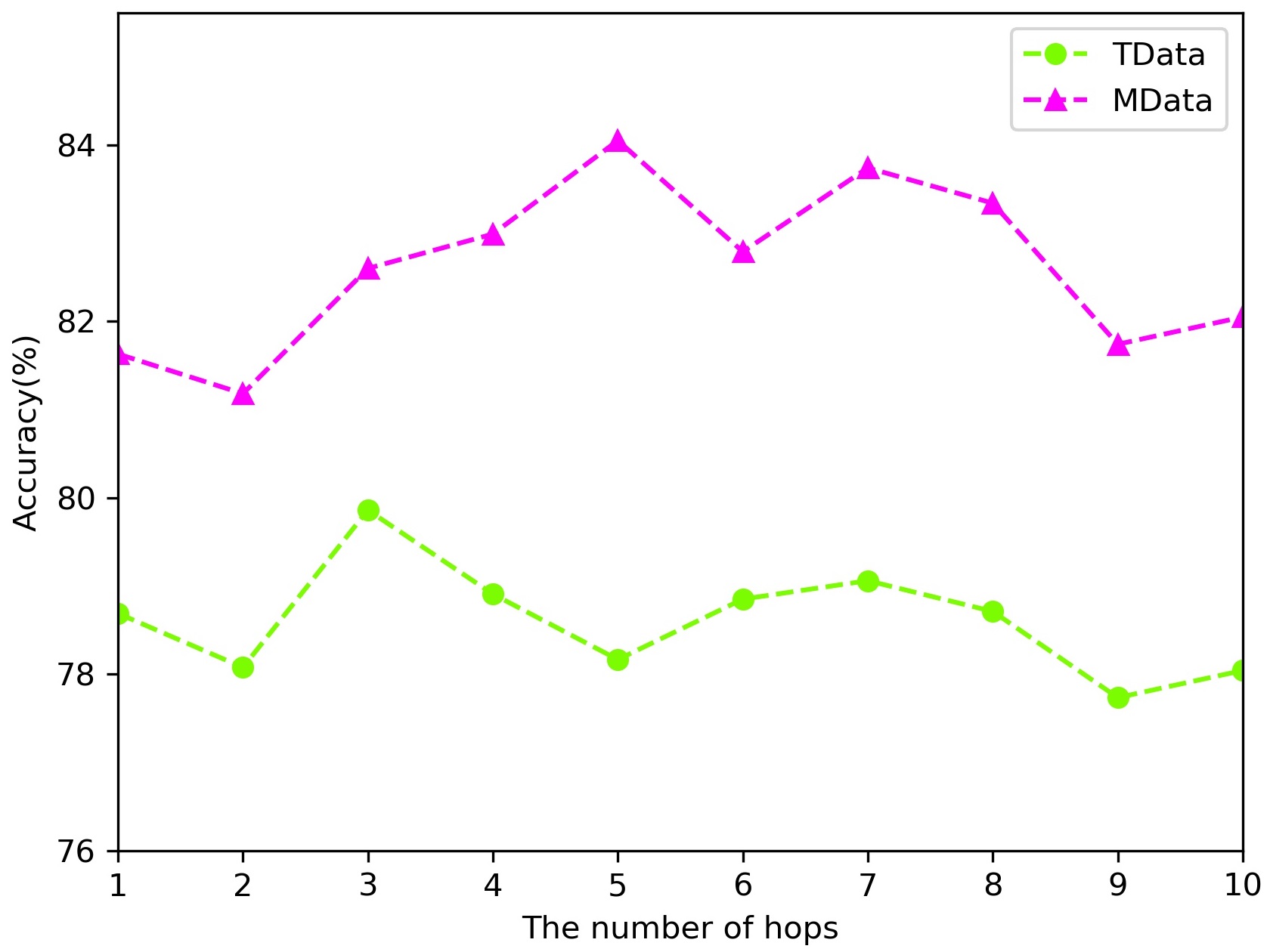}
	\caption{\label{Fig.6} Accuracies under the different numbers of hops for multi-hop based fusion.}
\end{figure}

In the multi-hop module, the number of hops is also an important parameter. We likewise perform experiments to explore the relationship between hop count and final accuracy. Fig. \ref{Fig.6} shows the accuracies of AntNet under different numbers of hops.\par
The number of hops also influences the final performance. The highest value (when the number equals 3) is nearly 2\% higher than the lowest value (when the number equals 9) on TData. On MData, the overall trend increases, and the accuracy is the highest when the number equals 5.

\subsection{Discussion}
We empirically analyze the error understanding answers in the test set to substantially scrutinize the performance of AntNet. The results show that errors are prone to occur for answers containing implicit preference information. In particular, once the implicit information contains negative or positive words, they are likely to be error judged. Table \ref{tab.6} shows several examples of answers containing implicit information. The first question belongs to the MC type. Hence, each label should correspond to an option term such as ``skirts" and ``pants".

The fourth question-answer pair is used as an example. The answer does not provide a direct reply to the question. In fact, the answer means that the user neither likes to run in the morning or night.
Future work will focus on the extraction of additional hints for users' choices.\par

Attention is the core of deep neural networks in NLP \cite{Vaswani2017}. The following example is visualized to facilitate the analysis of the effectiveness of the multi-hop attention used in this study.\par
\begin{figure}[htbp]
	\centering
	\includegraphics[width=0.8\textwidth]{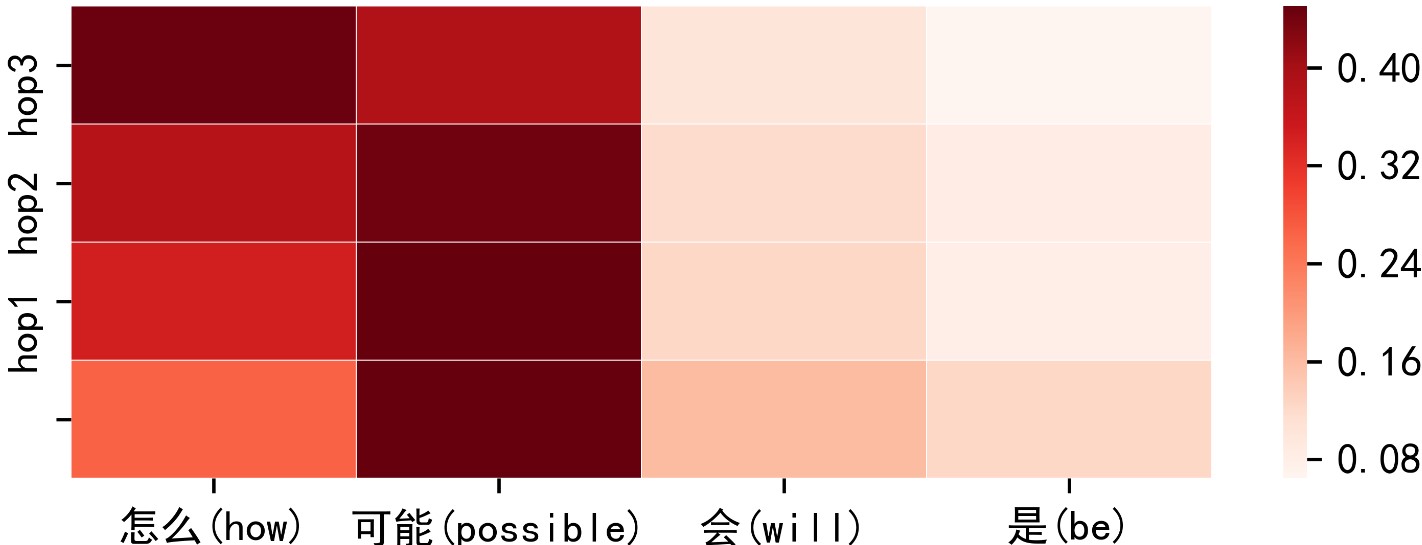}
	\caption{\label{Fig.7} Multi-hop attention scores for an answer sentence.}
\end{figure}

\begin{CJK*}{UTF8}{gkai}
In hop1 shown in Fig. \ref{Fig.7}, the attention score for the Chinese word ``怎么``(how) is small. Nevertheless, in hop4, its attention score becomes high, which is reasonable because the Chinese word is quite important for answer understanding.
\end{CJK*}

\begin{figure}[htbp]
	\centering
	\includegraphics[width=0.6\textwidth]{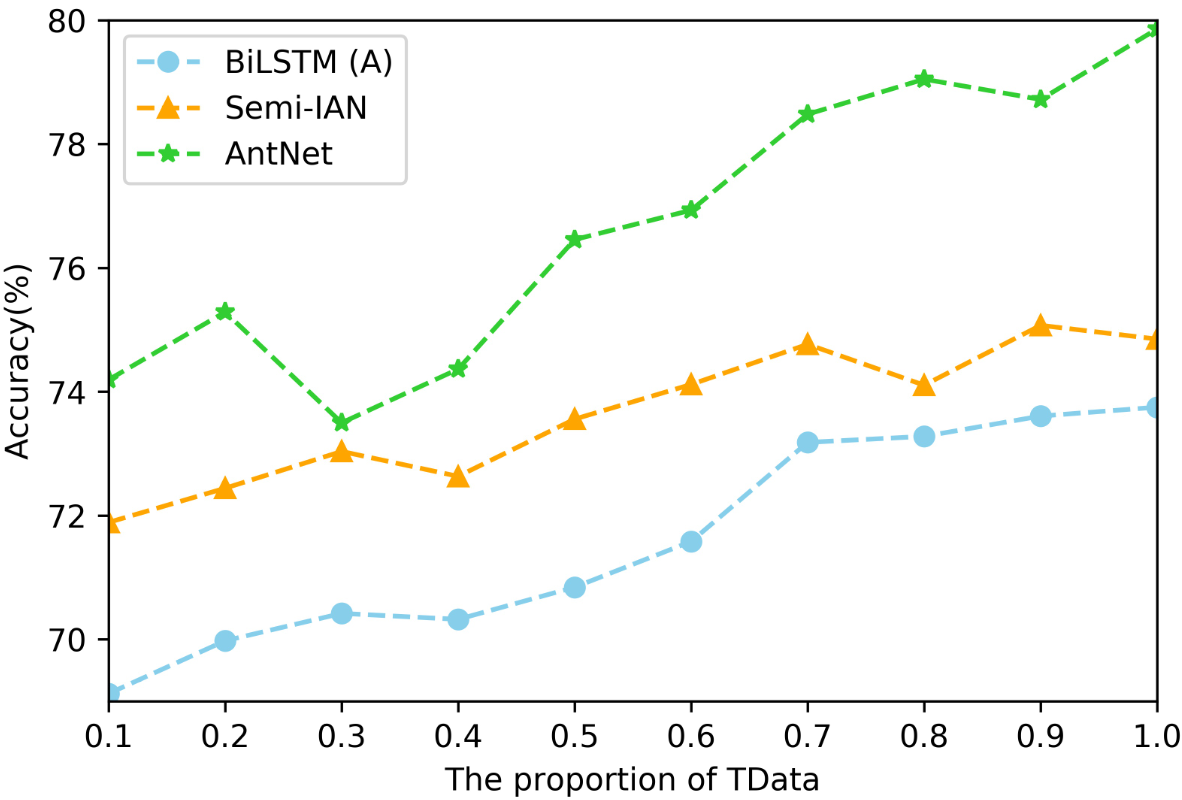}
	\caption{\label{Fig.12} Accuracies under the different proportions of training data on TData.}
\end{figure}

We also investigated the relationship between training data and model performances. Fig. \ref{Fig.12} shows the variations of performances under different proportions of training data on TData. With the increase of training data, the performances of the three methods, i.e., AntNet, BiLSTM, and semiIAN, also increased. Nevertheless, when the training data is small, the performance of AntNet is also relatively good. Similar observations are obtained on MData.

\section{Conclusion}
The automatic understanding of human answers in reverse QA can bring natural interactions, thereby improving user experiences. However, this topic receives little attention in the previous literature. The current study compiles a relatively large data corpus for answer understanding in reverse QA. An effective deep neural network called AntNet is proposed to understand the answers for the two most common types of questions.

AntNet utilizes two types of questions and a relevance-aware presentation for answer texts. The multi-hop based fusion module is used to model the contextual dependency between questions and answers. The experimental results indicate that AntNet is significantly better than the existing method and state-of-the-art NLP models with direct variations.

\begin{acks}
We thank Dr. Guan Luo, Dr. Xiaodong Zhu, and Prof. Qinghua Hu for their contributions in data collection and our early proposed model, Semi-IAN in PAKDD paper. We also appreciate the anonymous reviewers for their insightful comments and constructive suggestions.
\end{acks}

\bibliographystyle{ACM-Reference-Format}
\bibliography{main}

\end{document}